# IRANIAN CASHES RECOGNITION USING MOBILE

Ismail Nojavani [1], Azade Rezaeezade [2] and Amirhassan Monadjemi [3]

[1]Department of Computer Engineering, University of Isfahan, Isfahan, Iran
e.nojavani@eng.ui.ac.ir
[2] Department of Computer Engineering, University of Isfahan, Isfahan, Iran
a.rezaeezade@eng.ui.ac.ir
[3] Department of Computer Engineering, University of Isfahan, Isfahan, Iran
monadjemi@eng.ui.ac.ir

## ABSTRACT

*In economical societies of today, using cash is an inseparable aspect of human life. People use cashes for marketing, services, entertainments, bank operations and so on. This huge amount of contact with cash and the necessity of knowing the monetary value of it caused one of the most challenging problems for visually impaired people. In this paper we propose a mobile phone based approach to identify monetary value of a picture taken from cashes using some image processing and machine vision techniques. While the developed approach is very fast, it can recognize the value of cash by average accuracy of about 95% and can overcome different challenges like rotation, scaling, collision, illumination changes, perspective, and some others.*

## 1. INTRODUCTION

From the beginning of human appearance on the earth, we always gain lots of information about our surrounding environment visually. As result, our developed technologies mostly are based on vision. Hence, human beings who have some kind of visually impairment always have suffered from that. One of the most common innovations of humans that proof this assertion is money. Visually impaired people have lots of difficulties to use money in daily transactions, unless they use a person or device as an assistant to help them. World Health Organization approximates 285 million people are visually impaired worldwide, that 39 million of them are blind. According to the Health Organization of Iran, about 600 thousands Iranian have moderate or severe visual impairments that about a third of them are totally blind. This statistics show the importance of developing some efficient methods to assist this people in cash recognition. In this paper we develop a mobile phone based cash recognition system that helps visually impaired people to identify the current Iranian cash easily and accurately.

Banknotes in different countries have texture, colour, size, and appearance differences [1], so feature extraction and identification approaches that is used in one country usually is not usable in other countries or does not work properly over there. Moreover, lots of previous work in concept of cash recognition is restricted to specific standard conditions. For example many of the developed approaches do not support rotation or noisy backgrounds [2-5], or in some other approaches the whole banknote must be visible in taken picture [1, 6]. By considering visually impaired limitations to take pictures with special physical characteristics, these approaches are not user friendly enough.

In this study we consider many complexity and variety for taken pictures and as a result the developed system can support rotation, scaling, complex or noisy background, camera angel changes, collision and even the variation of illumination. For explication we define this concept below and show some of them in Figure 1.

- *Rotation:* the cash can have different rotation angle to the camera.
- *Scaling:* the cash can poses in different distances from the mobile camera.
- *Complex background:* the cash can be taken with any background except other cashes.
- *Perspective:* the cash can poses in any camera angle.
- *Collision:* the taken picture can be from a part of cash. It should only contain the digits of monetary value.
- *Variation of illumination:* cash picture can be taken in different illumination.

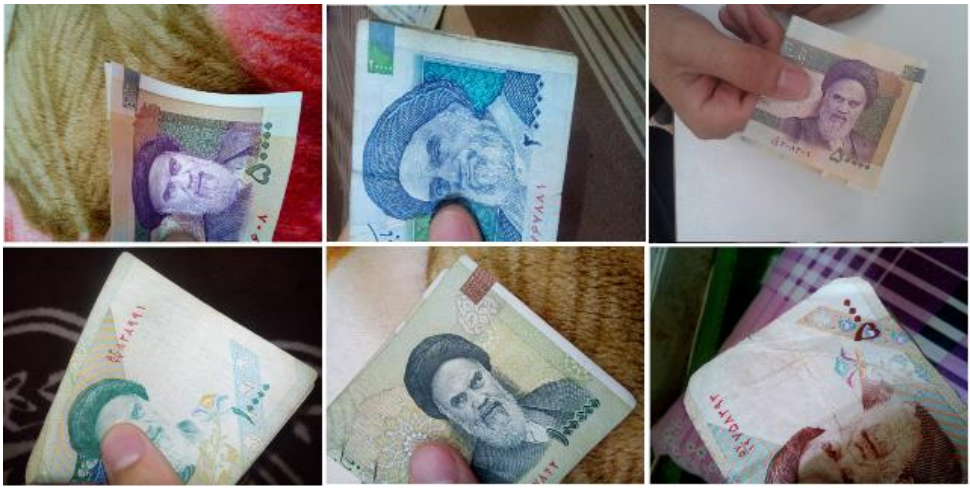

Figure 1. Some possible varieties in taking picture (oldness, rotation, scaling and different illumination)

In the developed approach, for robust and flexible cash recognition, we use value digits exist on cash. However, regarding the different conditions of taking pictures that was shown in Figure 1, these digits are not in a constant place for all the pictures. So, first of all we should find the place of these digits on the cash image. In this study, to localize the value digits on the image we apply a zero-finding algorithm on it. After that, we proceed to find the remained digit that can be 1, 2 or 5, and try to identify this digit using a neural network. Besides that, we should count the number of zeroes. After these two steps we can identify the monetary value of that cash.

The contributions of this paper include: section 2 reviews the state of the art on banknote recognition, section 3 explains the proposed approach for Iranian cash recognition, section 4 evaluates the recognition results and section 5 conclude the whole paper.

## 2. RELATED WORK

The existing cash recognition approaches in literature, mainly use image processing and neural network techniques. Moreover, there is some developed devices which use physical characteristics of cashes like size or colour. The most considerable point about cash recognition is that many developed devices or methods for cash recognition in one country are not usable in other countries. In this section we will introduce some of these devices and methods.

Money Talker [7] is a device that recognizes Australian bank notes electronically, using the reflection and transmission properties of light. This device uses the largely different colours and patterns on each Australian banknote. Different colour lights are transmitted through the inserted note and the corresponding sensors detect distinct ranges of values depending on the colour of the note. Cash Test [8] is another device determines the value of banknote by a mechanical means relying on the different lengths of each notes. The downfall of the Cash Test is that it does not allow for the shrinking of notes with time nor the creases or rips that are common in our notes. Moreover, while Cash Test is cheap and extremely portable by the reports of users, it is inaccurate and difficult to use. Kurzweil reader [9], iCare [10] (uses a wearable camera for imaging and a CPU chip for computation) and some other devices have been developed too, each has their advantages and disadvantages, but the common property of them is that the user most carry a device everywhere. Some of them are bulky and expensive too.

In [1], Hassanpour and Farahabadi proposed a paper note recognition method which uses Hidden Markov Model (HMM). This method models cashes using texture characteristics including size, colour and a texture-based feature from banknote and compares extracted vector with the instants in a data base of paper cashes. But the main purpose of this method is to distinguish national banknotes from different countries. In [11], Lui proposed a background subtraction and perspective correction algorithm using Ada-boost framework and trained it with 32 pairs of pixels to identify the bill values. This system uses video recording and works on snapshots. This method does not support rotation and noisy background and strongly relays on the white and straight edges of bill, so it is not usable for Iranian paper cash recognition because they do not have this required margins usually.

In [12] Hassanuzzaman and Yang proposed a component based framework using SURF features. The proposed method achieved 100% accuracy for US dollars through various conditions. While the proposed framework is a robust and effective approach and is available to overcome challenges like image rotation, noisy background, collision, scaling and illumination changes, it is not a suitable method for Iranian banknote recognition. This disproportion refers to the texture difference of Iranian currency tissues and US dollars. For proving this unfitness we have applied this approach to Iranian cashes and checked it by extracting different components. But the output accuracy was extremely low which has convinced us that SURF is not a suitable approach for Iranian cash recognition.

As we can see, while all these methods have their advantages, we cannot use them in a user friendly and accurate manner for Iranian cash recognition. So we are going to develop a user friendly method which is portable with no difficulties, have enough accuracy, and considering impaired people limitation for taking pictures in the next sections.

## 3. THE PROPOSED APPROACH

In this section we will explain the developed system completely. As mentioned earlier, in this method we should provide a picture of the cash that contain the monetary digits. The taken picture must be from the side that contains Iranian digits. For satisfying this condition, photo taker should use face detection technology of his mobile phone. In front side of all Iranian cashes, there is a face of Imam Khomeiny. So if the user is taking picture from the front side his mobile will detect Imam's face. But if he is taking picture from the back side his mobile would not detect any face, so he will understand that the cash must be turned back. We assume this picture is taken by a mobile phone camera, so it does not have high resolution and is taken by a visually impaired person, so can have any position. Then this picture will be analysed by machine vision and image processing tools to extract the individual features of the cash and to recognize the value of that cash. After that, the identified monetary value will be revealed to the user.

Feature extraction should be done by approaches that are robust to different environmental conditions that reviewed earlier including rotation, scaling, noisy background, collision, perspective, camera different angle, variance in illumination and any other possible situation. For this process we will have two main steps. The firs one is localization of monetary value digits. This step is carried out by means of some image processing methods and some innovation in interpreting appeared concepts. Second step is recognition of the monetary value of cash. In this step we will use neural network to identify the nonzero digit and identify the value of that cash. In the following we will explain these two steps.

The mentioned face detection process, localization step for finding zeroes and counting them and identifying nonzero digits are shone in Figure 2.

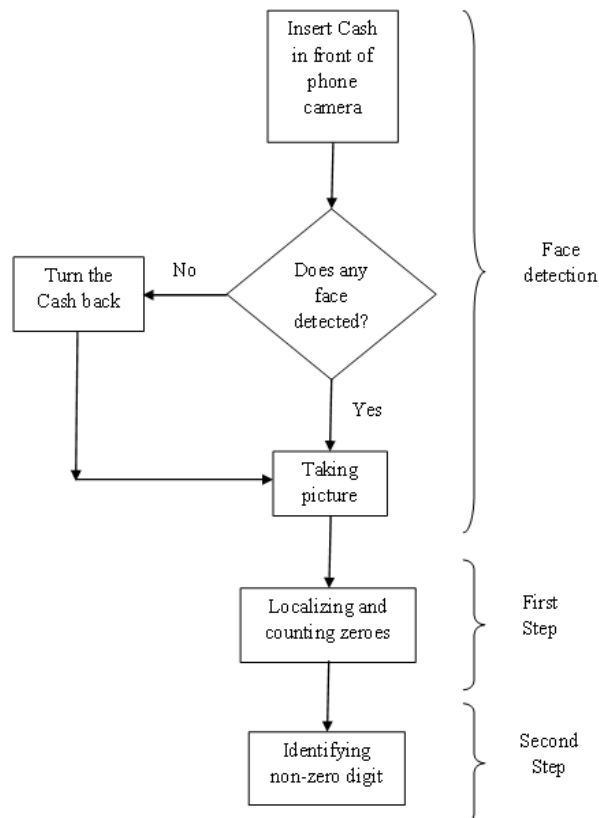

Figure 2. Summary of main steps in proposed approach

## 3.1. Localization of the Monetary Value

In the first step, to find the place of monetary value digits, the RGB image of the cash should be changed to a grayscale one; it is because all the next processes for feature extraction and cash identification are supposed to be done on a mobile device, so it should not be time and CPU consuming. We apply a Wiener filter on the grayscale image to reduce the effect of oldness of cash or some noises on the image. This grayscale image then will be converted to a binary image. Since photo taker can use any mobile phone camera with any quality, and regarding to that the user can take picture in variant illumination and with different background, for conversion of grayscale image to a valid binary one we use a local adaptive approach for determining the optimum threshold. The used approach is explained in [13]. In Figure 3(a) a sample of original picture taken by a mobile phone camera is shown. The binary image after grayscale conversion and using local adaptive threshold is shown in Figure 3(b). As you see, in this picture we have a lot of black and white regions. We refer to these white regions as components.

In some Iranian cashes, like 20000, 50000 and 100000 Riyals, because of existence of some huge white components near the zeroes, there is a possibility that after binarization, these zeroes and big components stick to each other. Moreover, we have this problem for zeroes in 100000 Riyals. For example in Figure 3(c), a bigger region belongs to Imam's face is stuck to zeroes. So, for separation of these two regions, we use a 3×3 median filter to delete some small components on image and to fill some small holes created on zeroes.

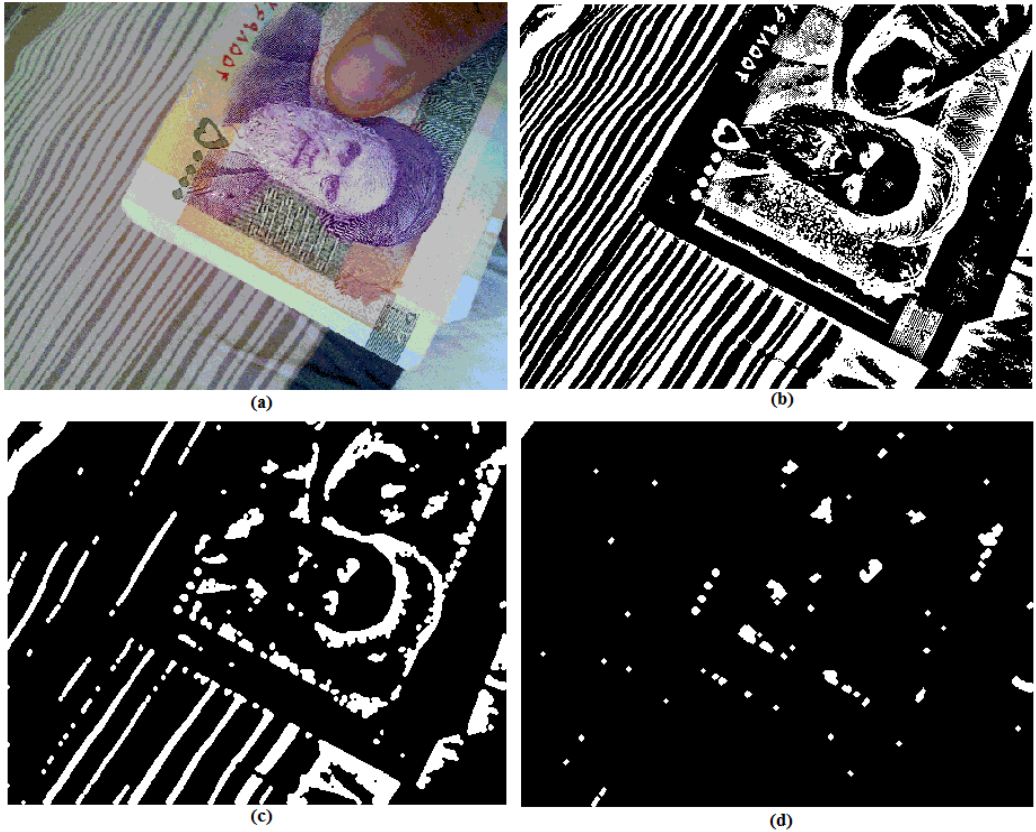

Figure 3. (a) A sample of original taken picture, (b) Binary image after applying local adaptive threshold (c) Corresponding image after applying median filter, erosion and dilation operators. (d) Corresponding image after elimination of regions that have a length to width ratio greater than 2 or number of white pixel to boundary rectangle area pixel ratio greater than two.

In continuous, we are going to eliminate regions which cannot be a batch of zeroes because of their shapes. In Persian orthography the width and length of zeroes must be approximately the same. It means that the width and length of boundary rectangle of zeroes on cash or on the taken picture must be approximately equal. So we can omit regions which the ratio of their boundary rectangle length to its width is more than 2. So, components in a region with too different width and length can be omitted with a high reliability. Moreover, while the shape of component are irregular, there are some component that the ratio of length and width of their boundary rectangle is less than 2, but if we calculate the ratio of their white pixel to the boundary rectangle area it is greater than 2. It means that these components cannot be zero too. So we can omit them again. The resulted image is shown in Figure 3(d).

While the existing zeroes in the image of cash are near to each other, if we apply closing morphology operator [14] to the image, these zeroes and some other components will stick to each other again. The result image is shown in Figure 4(a).

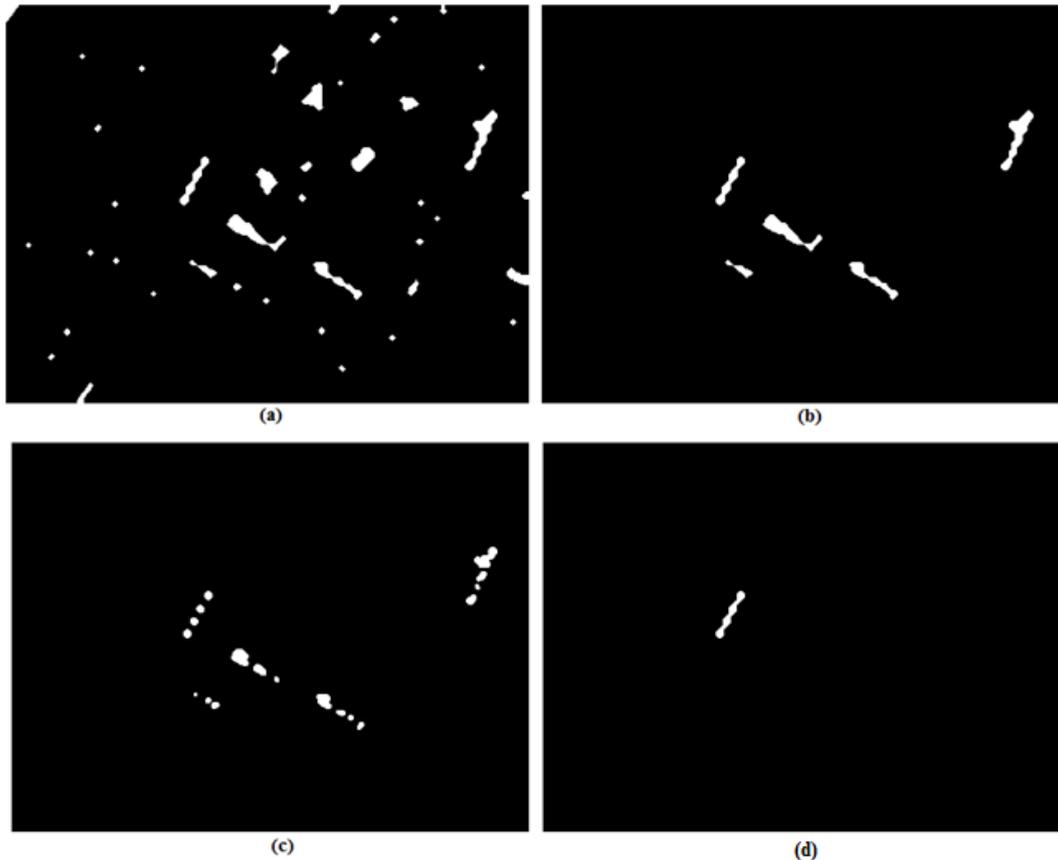

Figure 4. (a) Corresponding image after using closing operator. (b) Elimination of regions with less than 3 components. (c) Corresponding image after elimination of regions and components which are not belong to an acceptable line. (d) Keeping the region with the least difference between its components.

The lowest number of zeroes on current Iranian cashes in is 3. So after applying closing operator, we can delete connected regions which are made up of less than 3 components. This will limit the number of questioned regions for being a batch of zeroes seriously. Resulted image is shown in Figure 4(b).We should notice that it is not correct to omit regions with more than 5 components (5 is the number of zeroes in 100000 Riyals) because some nonzero components might be connected to the zeroes too.

According to the existence of zeroes on a unique line, we can find regions with more than 2 components on a communal line and omit other regions and then in remained regions omit components which are not on the corresponding line. For this purpose, we extract a line in each connected region crossing gravity centres of any two components of that region. For example, if we have a region with 3 connected components, we should establish equation for 2 lines. The first one must cross gravity centre of the first and second components. The second one must cross gravity centre of the second and third components. After that, for each line we should test if it crosses other components of the connected regions or not. In the previous example, firstly we should test that the first line crosses the third component or not and then we should check if the second line crosses the first component or not. Notice that in this step existence of only one pixel of the component on that line is sufficient. For each connected regions we keep the line that have the most components on it. In addition, for each connected regions any other components which are not on the established line should be omitted because they cannot be zero. The result is shown in Figure 4(c).

As it is shown in Figure 4(c), it is possible that the algorithm keeps more than one line. Now we should expand a solution to find the right region that contains zeroes. For this purpose we notice the size of zeroes that should approximately be the same. So, for each region we get the size of all components. After comparing the size of all components in each region, we keep only the region that its components have the least difference and omit the other lines and regions. The result is shown in Figure 4(d).

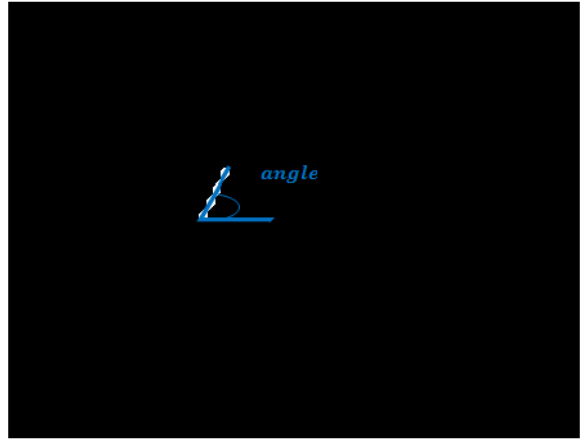

Figure 5. Obtaining the rotation angle

Surprisingly we can compute the rotation angle of the cash by means of corresponding line of the last remained region which should be the batch of zeroes. It is just enough to obtain the angle of the line that crosses the centre of first and last zeroes as it is shown in Figure 5. We can compute the angle from (1).

$$angle = arctang\left(\frac{y_1 - y_0}{x_1 - x_0}\right) \quad (1)$$

In this equation, $y_0$ is central width of first zero, $y_1$ is central width of last zero, $x_0$ is central length of first zero and $x_1$ is central length of last zero. Now if we rotate the image of Figure 3(b) by angle, the rotated image in Figure 6 will be generated. Then, we can easily count the number of zeroes. This count will be used for identifying the monetary value of image.

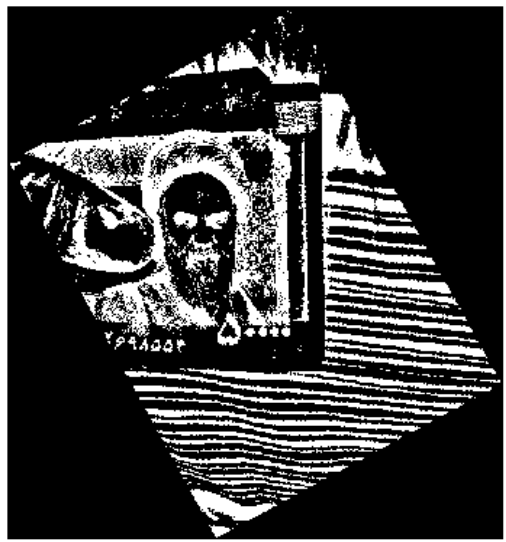

Figure 6. Corresponding image after rotation by angle

## 2.2. Recognition of Monetary Value

Till now, we could find place of zeroes and the rotation angle. Now we can fine the nonzero digit using place of zeroes. In this section we will discuss the features extracted for remained digit. To find the nonzero digit, we start moving to right and left from the most right zero and the most left one equally. We will stop when we find the first component. This component certainly is our nonzero digit. For extracting this digit from the image, we use the length of crossed line from first zero to the last. In all cashes the nonzero digit will be fixed in a square by size of L that L is obtained from (2):

$$L = \frac{3 * length\ of\ line}{number\ of\ zeroes} \qquad (2)$$

One of the vertical sides of square is very close to last zero and the centre of this side is intersecting by the corresponding line crosses zeroes. The other vertical side of square is parallel with the first one and has distance of size L from it. According to this we can extract the nonzero digit of monetary value. Figure 7 shows an extracted digit using this approach. Probably we will have some small components in the extracted square but we remove them by keeping the largest connected component and delete remainders. So we will maintain only the nonzero digit.

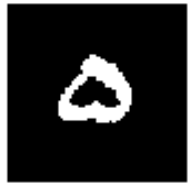

Figure 7. Extracted nonzero digit

Now it is time to identify the number exist in extracted square. As we mentioned earlier, remained digit always is 1, 2 or 5. So for identification of any digit remained after finding it, we must extract features that can classify these digits accurately, so we choose features as below:

1. Ratio of digit pixels to boundary rectangle pixels.
2. Ratio of height of digit to its width (it is computed as ratio of length to height of boundary rectangle for each digit)
3. Ratio of left length to right length and ratio of top width to bottom width.
4. Horizontal and Vertical Symmetry that compare corresponding features in top and down half and right and left half.

For classification we used a Multi-Layer Perceptron (MLP) neural network with one hidden layer. For obtaining the best number of neurons in hidden layer, the network has been trained and tested with different number of hidden nodes and the best result has been achieved by 20 neurons in one hidden layer.

## 4. EVALUATION OF THE PROPOSED METHOD

In this section we will demonstrate the accuracy and performance of our method. As we mentioned earlier, proposed approach is a two-phase method:

1. Localization of monetary digits and counting the zeroes
2. Identifying nonzero digit of the monetary value

So for computing the accuracy of this method we must use conditional probability. It is because of dependence of second phase accuracy to the first phase. We suppose that the accuracy of correct localization and counting number of zeroes is a% and accuracy of identifying remained digit correctly is b%, so total accuracy of proposed method can be compute via the conditional probability of (3):

$$accuracy = a\% * b\% \qquad (3)$$

For testing proposed approach we collect a data set containing 3500 image of current Iranian cashes in different states and by a vast amount of variety (different environmental conditions we mentioned in the introduction). We took these pictures by different mobile phone cameras which have different resolution from 3 to 8 mega pixels. The rate of accurate localization of zeroes and accurate count of them (first phase) for different cashes are shown in Table 1.

Table 1. Accuracy of localizing monetary value
and counting number of zeroes

| Cash | Accuracy |
| --- | --- |
| 1000 Riyals | %96.8 |
| 2000 Riyals | %91.6 |
| 5000 Riyals | %96 |
| 10000 Riyals | %98.4 |
| 20000 Riyals | %93.6 |
| 50000 Riyals | %95.2 |
| 100000 Riyals | %89.4 |
| **Average** | 94.43% |

As we mentioned earlier in 100000 Riyals cashes, the zeroes are near each other. Even in some cases they are stick to each other. Moreover, as you can see in Figure 8 in the case of 100000 Riyals cashes, there is an edge around zeroes. After binarization, this edge creates some noises around the zeroes that cause to incorrect localization of zeroes. The decrease in detection of zeroes place in the case of 100000 Riyals in Table 1 is because of this issues.

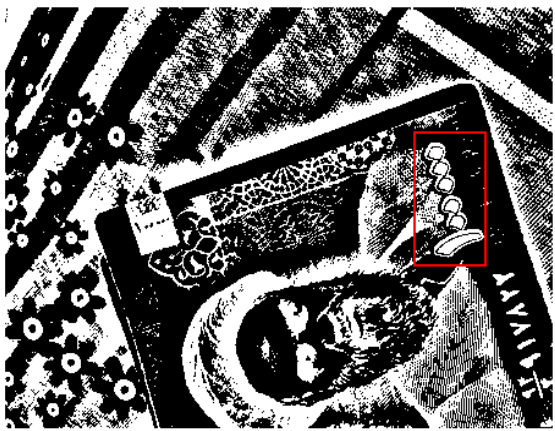

Figure 8: 100000 Riyal's zeroes has stuck to each other

As we mentioned earlier, we used MLP neural network with one hidden layer for identifying nonzero digit of monetary value. Number of neurons in the hidden layer is 20. Number of neurons in the input layer is 4 (the number of extracted features) and the number of neurons in the output layer is 3 (3 output including 1(١), 2(٢) and 5(٥) and we consider one neurons for each output). Transfer function for hidden layer is 'logistic' too.

Table 2. Accuracy of nonzero digits identification

| Nonzero Digit | Accuracy in Training Set | Accuracy in Test Set |
|---|---|---|
| 1(١) | 99.40% | 98.83% |
| 2(٢) | 99.69% | 99.28% |
| 5(٥) | 100% | 100% |
| **Average** | 99.7% | 99.37% |

For training and test of neural network, we divide created data set into 2 distinct data sets randomly. 70% of samples should be in training data set and reminded 30% should be in test data set. Rate of correct identification or accuracy in this phase is shown in Table 2.

Now by means of Table 1 and Table 2 and by using multiplying rule in equation (3), total accuracy will be obtained as what is shown in Table 3.

Table 3. Total accuracy for Iranian cash recognition

| Cash | Total Accuracy |
|---|---|
| 1000 Riyals | 95.67% |
| 2000 Riyals | 90.94% |
| 5000 Riyals | 96% |
| 10000 Riyals | 97.25% |
| 20000 Riyals | 92.93% |
| 50000 Riyals | 95.2% |
| 100000 Riyals | 88.35% |
| **Average** | 93.76% |

## 5. CONCLUSIONS

In this paper we presented a fast and accurate method for recognizing Iranian monetary value of cashes. This proposed method is based on image processing and machine learning approaches. As input we need a picture taken by mobile phone camera. So the resolution can be very low. We take into account the limitation of visually impaired people for taking pictures. So this proposed approach is robust to rotation of input picture, different scale of the taken picture, different perspective of camera, variation of illumination, noisy background and collision. Moreover, this method is robust to different design of cashes. All this robustness is because of special component we focus on; all the processes in this method are on monetary value of cashes.

In the first step of our method we fine the place of monetary value on the cash. This is down with a string of image processing procedures and operators. After finding the place of monetary value or more specially the place of zeroes we count the number of them, discover the rotation angle of input picture and extract the nonzero digit of monetary value. In second step we identify the nonzero digit by means of a MLP neural network with one hidden layer and 20 neurons in that layer. Outputs of this neural network are 1, 2 or 5 digits based on the input image. Based of accuracy of this step we could achieve total accuracy around 95%.

While the accuracy of this work is high in the field of Iranian cash recognition, and while it is the only implemented method of Iranian cash recognition for visually impaired people, the accuracy is not enough for a reliable detection. So one of the most important planes for future work is increasing the accuracy by means of machine vision approaches or testing other methods for nonzero digit identification.

## ACKNOWLEDGEMENTS

The authors would like to thank Dr. Sayed Amir Hassan Monadjemi, all the survey and study participant.

## REFERENCES

[1] Hassanpour, H., & Farahabadi, P. (2009) "Using Hidden Markov Models for paper currency recognition," *Expert Systems with Applications*, Vol. 36, No. 6, pp 10105-10111.

[2] Frosini, A., Gori, M. & Priami, P. (1996) "A neural network-based model for paper currency recognition and verification," *IEEE Transactions on Neural Networks*, Vol. 7, No. 6, pp 1482-1490.

[3] Kosaka, T. & Omatu, S. (1999) "Bill money classification by competitive learning," *in Proceedings of IEEE Midnight-Sun Workshop on Soft Computing Methods in Industrial Applications*.

[4] Kosaka, T., Omatu, S. & Fujinaka, T. (2001) "Bill classification by using the LVQ method," *in Proceedings of IEEE International Conference on Systems, Man, and Cybernetics*.

[5] Lee, J., Jeon, S. & Kim, I. (2004) "Distinctive point extraction and recognition algorithm for various kinds of euro banknotes," *International Journal of Control Automation and Systems*, Vol. 2, No. 2, pp 201-206.

[6] Jahangir, N., & Chowdhury, A. (2007) "Bangladeshi banknote recognition by neural network with axis symmetrical masks," *in Proceedings of IEEE 10th International Conference on Computer and Information Technology*, pp 1-5.

[7] Hinwood, A., Preston, P., Suaning, G. & Lovell, N. H. (2006), "Banknote recognition for the vision impared," *in Proceedings of Australasian Physical Engineering Sciences in Medicine*, Vol. 29, No. 2, pp 229-233.

[8] Reserve Bank of Australia, "How the RBA assists people with a vision impairment to differentiate notes," http://www.rba.gov.au/CurrencyNotes/vision_impaired.html, accessed 7/6/2010.

[9] K-NFB reader website, Available from: http://www.knfbreader.com/, accessed: 11/18/2014.

[10] Krishna, S., Little, G., Black, J. & Panchanathan, S. (2005) "A wearable face recognition system for individuals with visual impairments," *in Proceedings of the 7th internationa ACM SIGACCESS conference on Computers & accessibility*, pp 106–113.

[11] Liu, X. (2008 ) "A camera phone based currency reader for the visually impaired," in *ACM Proc. ASSETS'08*, Halifax, Nova Scotia, Canada, pp. 305-306.

[12] Hasanuzzaman, F. M., Yang X. & Tian, Y. (2009), "Robust and Effective Component-based Banknote Recognition by SURF Features," *in 20th Annual Wireless and Optical Communications Conference (WOCC)*, pp 10105–10111.

[13] Singh, T. R., Sudipta R., Singh, O. I., Sinam, T. & Singh, Kh. M. (2011) "A New Local Adaptive Thresholding Technique in Binarization," *IJCSI International Journal of Computer Science Issues*, Vol. 8, Issue 6, No 2.

[14] Serra, J. (1986) "Introduction to Mathematical Morphology," *Computer Vision, Graphics and Image Processing*, Vol. 35, No. 3, pp 283-305.

**Authors**

I. Nojavani born in 1987 in Bonab, Iran. He received his B.Sc. in computer software from Urmia University, Urmia, Iran, in 2010. Currently he is a M.Sc. student in Artificial Intelligence at the Department of Computer Engineering, University of Isfahan, Isfahan, Iran. His research interest contains OCR, DIP, and Artificial Intelligence.

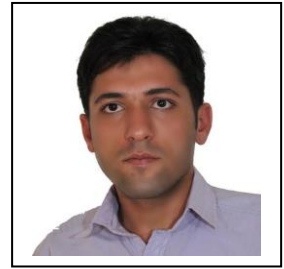

S. A. Monadjemi is born in 1968 in Isfahan, Iran. He received his B.Sc. in electrical/ computer engineering from Isfahan University of Technology, Isfahan, Iran in 1992, and his M.Sc. in computer engineering, machine intelligence from Shiraz University, Shiraz, Iran in 1995 and his PhD in computer engineering, processing and pattern recognition, from Bristol University, Bristol, England in 2004. His research interests are DIP, Machine Vision, Pattern Recognition, Artificial Intelligence, and Training through Computer. Dr. Monadjemi is currently an Asst. Professor in Department of Computer Engineering, University of Isfahan, Isfahan, Iran.

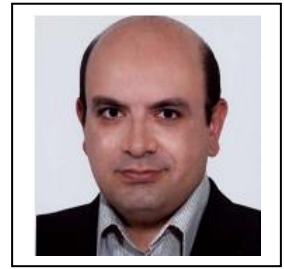

A. Rezaeezade born in 1988 in Aligudarz, Iran. She received her B.Sc. in computer hardware from Isfahan University of Technology, Isfahan, Iran, in 2010. Currently she is a M.Sc. student in Artificial Intelligence at the Department of Computer Engineering, University of Isfahan, Isfahan, Iran. Her research interest contains OCR, DIP, and Artificial Intelligence.

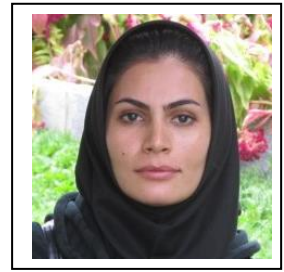